\pdfoutput=1

\documentclass[11pt]{article}

\usepackage[]{EMNLP2023}

\usepackage{times}
\usepackage{latexsym}
\usepackage{amsmath,calc}
\usepackage{graphicx}
\usepackage{multirow}
\usepackage{placeins}
\usepackage{arydshln}
\usepackage{float}
\usepackage{booktabs} 
\usepackage{subcaption}
\usepackage[normalem]{ulem}
\useunder{\uline}{\ul}{}
\usepackage[T1]{fontenc}

\usepackage[utf8]{inputenc}

\usepackage{microtype}

\usepackage{inconsolata}

\usepackage{xcolor}
\usepackage{CJKutf8}
\newcommand{\chinese}[1]{{\begin{CJK*}{UTF8}{gkai} #1 \end{CJK*}}}
\definecolor{bblue}{HTML}{4F81BD}
\definecolor{rred}{HTML}{c4260b}
\definecolor{ggreen}{HTML}{098c1f}

\DeclareRobustCommand{\hlred}[1]{{\textcolor{rred}{#1}}}
\DeclareRobustCommand{\hlblue}[1]{{\textcolor{bblue}{#1}}}
\DeclareRobustCommand{\hlgreen}[1]{{\textcolor{ggreen}{#1}}}

\usepackage{booktabs}
\usepackage{makecell}
\usepackage{arydshln}

\title{Towards Effective Disambiguation for Machine Translation \\ with Large Language Models}

\author{
Vivek Iyer \quad Pinzhen Chen \quad Alexandra Birch
\\
School of Informatics, University of Edinburgh\\
 \texttt{\{vivek.iyer, pinzhen.chen, a.birch\}@ed.ac.uk}\\
}

\begin{document}
\maketitle
\begin{abstract}

Resolving semantic ambiguity has long been recognised as a central challenge in the field of Machine Translation. Recent work on benchmarking translation performance on ambiguous sentences has exposed the limitations of conventional Neural Machine Translation (NMT) systems, which fail to handle many such cases. Large language models (LLMs) have emerged as a promising alternative, demonstrating comparable performance to traditional NMT models while introducing new paradigms for controlling the target outputs. In this paper, we study the capabilities of LLMs to translate ``ambiguous sentences" - i.e. those containing highly polysemous words and/or rare word senses. We also propose two ways to improve their disambiguation capabilities, through a) in-context learning and b) fine-tuning on carefully curated ambiguous datasets. Experiments show that our methods can match or outperform state-of-the-art systems such as DeepL and NLLB in four out of five language directions. Our research provides valuable insights into effectively adapting LLMs to become better disambiguators during Machine Translation. We release our curated disambiguation corpora and resources at \url{https://data.statmt.org/ambiguous-europarl}.

\end{abstract}

\section{Introduction}
\label{sec:introduction}
While the field of NMT has advanced rapidly in recent times, the disambiguation and translation of ambiguous words still remain an open challenge. Notably, \citet{campolungo-etal-2022-dibimt} created a benchmark named DiBiMT to study the behaviour of state-of-the-art (SOTA) NMT systems when translating sentences with ambiguous words.\footnote{\url{https://nlp.uniroma1.it/dibimt/public/leaderboard}} They reported that even the best-performing commercial NMT systems yielded accurate translations only 50-60\% of the time,\footnote{Subsequent iterations of these commercial models have improved, but large margins still remain.} while other open-source multilingual models like mBART50 \citep{tang-etal-2021-multilingual} and M2M100 \citep{fan2021beyond} performed much worse. This was found to be due to biases against rare and polysemous word senses inherited during pretraining. Table~\ref{tab:eye-catcher} shows an example from the DiBiMT benchmark where DeepL\footnote{\url{https://deepl.com/en/translator}} mistranslates an ambiguous word while the LLM BLOOMZ resolves the word to its correct in-context meaning.

\begin{table}[t]
    \centering\small
    \setlength{\tabcolsep}{0.25ex}
    \begin{tabular}{ll}
        \toprule
        Source & The horse had a {\hlblue{blaze}} between its eyes. \\
        \midrule
        DeepL & \chinese{那匹马的两眼之间有一团\hlred{火焰}。} \\
        & (There is a {\hlred{flame}} between the horse's eyes.) \\
        \midrule
        \multirow{2}{*}{\makecell{BLOOMZ\\(176B)}} & \chinese{这匹马的眼睛之间有一道\hlgreen{白线}。} \\
        & (There is a {\hlgreen{white line}} between the horse's eyes.) \\ 
        \bottomrule
    \end{tabular}
    \caption{An example of English-to-Chinese translation involving an ambiguous term ``blaze''. For BLOOMZ, we use 1-shot prompting to obtain the translation.}
    \label{tab:eye-catcher}
\end{table}

In this paper, we explore whether LLMs can indeed perform better at translating ``ambiguous sentences" -- i.e. those containing highly polysemous and/or rare word senses. The motivation behind this is that while NMT models can potentially learn biases from noisy or narrow domain parallel data, hurting their ability to detect and translate rare word senses, LLMs can potentially be pretrained on a wider variety of monolingual text -- though they might also prefer fluency over accuracy. Still, LLMs have shown many emergent abilities due to scale \citep{brown2020language,chowdhery2022palm,wei2022emergent} and moreover, have demonstrated great potential for Machine Translation (MT) \citep{vilar-etal-2023-prompting, zhang2023prompting}.

We comprehensively examine how these trends extend to the specific task of translating ambiguous sentences. We select a diverse set of foundational and instruction-tuned LLMs, of different sizes and with varying combinations of languages in the pre-training data. We then compare how these LLMs match up against several widely used NMT models on the DiBiMT test set, which covers translation from English to five languages: Spanish, Italian, German, Russian and Chinese.  We find that, with only 1-shot in-context learning \citep{brown2020language},  LLMs -- in particular, BLOOMZ 176B \citep{muennighoff-etal-2023-crosslingual} and LLaMA 65B \citep{touvron2023llama} -- match or outperform top-performing open-source and commercial MT systems, and set a new SOTA in two of the five languages we tested. Furthermore, we propose two methods for adapting LLMs for ambiguous translation: 1) in-context learning with sentences having the same word sense, and 2) fine-tuning on curated ambiguous parallel corpora. We show that these methods are highly effective and can further improve performance by up to 15 points in DiBiMT accuracy in the best case.  

Our work thus makes three key contributions:
\begin{enumerate}
    \item We evaluate the performance of LLMs compared to top-performing NMT systems in the challenging task of translating ambiguous sentences. We report SOTA scores on 2 of the 5 languages tested, and comparable performance otherwise.

    \item We also show that our suggested techniques of similar sentence in-context learning and targeted disambiguation fine-tuning significantly outperform naive few-shot prompting

    \item We conclude our work by evaluating LLMs on the FLORES200 test sets, and confirm that improvements in disambiguation accuracy correlate strongly with those in overall MT quality.
\end{enumerate}

\section{Background}
\label{sec:background}
\subsection{Ambiguity in machine translation}

Resolving ambiguity in the source sentence was historically framed as one of the most fundamental challenges in MT \citep{weaver1952translation}. In an effort to address this challenge, traditional works integrating Word Sense Disambiguation in Statistical Machine Translation \citep{carpuat-wu-2007-improving, chan2007word} were followed by those integrating it in NMT architectures in various ad-hoc ways \citep{CHOI2017149, liu-etal-2018-handling, pu2018integrating}. Later, with the introduction of the Transformer \citep{vaswani2017attention}, it was shown that higher layer encoder representations are robust enough to handle disambiguation \citep{tang-etal-2019-encoders} without any explicit handling of word senses. 
    
However, more recent research creating challenging evaluation benchmarks has called the purported abilities of NMT systems into question once again. Following the proposal of the MuCoW benchmark for testing WMT19 \citep{raganato2019mucow} and WMT20 \citep{scherrer-etal-2020-mucow} systems, \citet{raganato2020evaluation} showed how Transformer-based NMT models, in general, underperform when translating rare word senses. \citet{campolungo-etal-2022-dibimt}, who experimented with SOTA commercial (Google Translate, DeepL) and open-source systems (mBART50, M2M100, OPUS-NMT \citep{tiedemann2020opus}, etc.), arrived at the same conclusion when they proposed the DiBiMT benchmark for evaluating MT systems between English and 5 languages (Spanish, Italian, German, Russian, and Chinese). They found similar biases against low-frequency and highly polysemous word senses. They also noted the accuracies of these systems were much lower than the then SOTA WSD system, ESCHER \citep{barba2021esc} -- indicating significant room for improvement. In this work, we explored whether foundational and instruction-tuned LLMs could bridge this gap with minimal supervision (i.e. few-shot prompting).
    
\subsection{LLMs and machine translation}

Previous research has found that LLMs can perform machine translation without being specifically fine-tuned \citep{radford2019language}. In order to elicit a translation, research in this direction follows the paradigm of LLM prompting:
\begin{enumerate}
    \item Zero-shot prompting, where an LLM is directly asked to translate a source input into the target language \cite{radford2019language}.
    \item Few-shot prompting, also called in-context learning, where an LLM is supplied with demonstrations of input and output pairs from the same task it is performing, before being queried an input \cite{brown2020language}.
    \item Chain-of-thought (CoT), where an LLM is prompted to reason to gain relevant knowledge about the input before producing an output \citep{chain-of-thought, kojima2022large}.
\end{enumerate}
Besides training-free approaches, another route is instruction tuning, which optimizes an LLM on a mixed range of downstream tasks and fine-tunes the model to understand and respond to user intention through natural language \cite{wei2021finetuned}.

\begin{figure*}[t]
\centering 
  \includegraphics[width=0.9\textwidth]{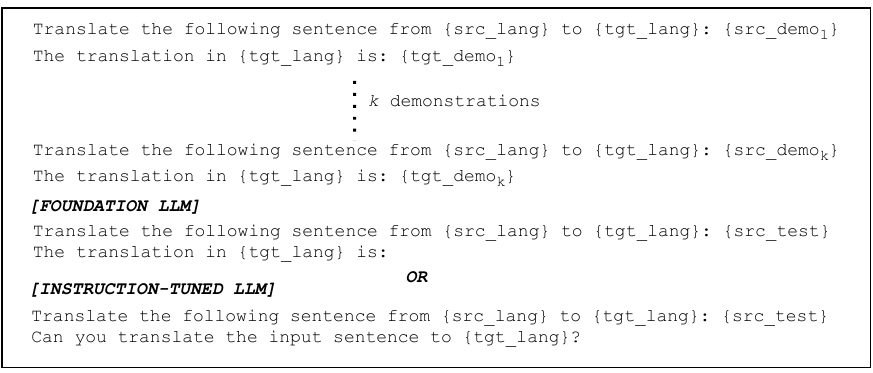}
  \vspace{-1ex}
    \caption{Templates used for $k$-shot LLM prompting, with $k>=0$.}
    \label{fig:PromptTemplate}
\end{figure*}

It was observed that LLMs might not surpass Transformer models solely trained to translate, especially for non-English and low-resource translation directions \citep{vilar-etal-2023-prompting,Hendy2023HowGA}. Nevertheless, LLMs have been shown to achieve superiority in tasks requiring in-depth understanding and manipulation of text, primarily due to them being pretrained on very large corpora. For example, without fine-tuning, LLMs are good at adapting to word alignments \citep{Moslem2023AdaptiveMT}, translation evaluation \citep{kocmi2023large}, idiom translation \citep{raunak-etal-2023-gpts},  iterative refinement \cite{chen2023iterative}, and interactive translation via CoT \cite{pilault2023interactive,he2023exploring}. Related to our work is \citet{pilault2023interactive}'s proposal of using interactive question answering as a CoT process for LLMs to disambiguate source words. As an alternative approach, we aim to generate translations in a single pass by leveraging SOTA WSD systems to provide contexts that guide LLMs to disambiguate better.

\section{Methodology}
\label{sec:methodology}

  
    
    
    
    

\subsection{Preliminaries}

A word sense is a concept in a Knowledge Base (in this work, BabelNet by \citet{navigli2021ten}) that denotes a distinct meaning of a word in the context of a sentence. The polysemy degree of an ambiguous word is defined as the total count of all possible senses that a particular word can have. The sense frequency is defined as the occurrence count of that particular sense in a disambiguated training corpus.

In this work, we define an ambiguous word as a polysemous term with multiple possible, and likely related, meanings -- with the correct sense inferable only from the sentence-level context. We then refer to a sentence with an ambiguous word as an ``ambiguous sentence'' for brevity and ease of explanation. By definition, the DiBiMT test set \citep{campolungo-etal-2022-dibimt} contains only one ambiguous word per sentence.

Word Sense Disambiguation (WSD) is the process of linking an ambiguous word in a sentence to its appropriate word sense in the Knowledge Base. We use ESCHER-WSD \citep{barba2021esc} in this work, a high-performing WSD system that had achieved the SOTA for English.

\subsection{\textit{K}-shot prompting}
Given a test sentence $X$ and a Large Language Model to prompt for translations, we construct a query with $k$ demonstrations, i.e. parallel sentence pairs $\{(X_1, Y_1), (X_2, Y_2) \ldots (X_k, Y_k) \}$ as examples, followed by the test sentence. As shown in Figure \ref{fig:PromptTemplate}, for foundation LLMs, we frame the prompt as a text completion task, while for instruction-tuned LLMs (like BLOOMZ) we structure the last phrase as a question, in order to conform to the latter's question answering format. In the naive setting, we choose our demonstrations randomly from the development set.

\subsection{In-context learning with similar ambiguous contexts}
\label{sec:simsentence-prompting}

LLMs can effectively gain knowledge relevant to the test domain through prompting, and this process is named in-context learning (ICL). We leverage ICL to help LLMs ingest information on translation of ambiguous sentences, by providing related sense translations as examples in the prompt. To achieve this, we first identify the most polysemous word in the input sentence by disambiguating it with a WSD system, and then calculate the polysemy degree of all disambiguated senses with respect to a large development set. We choose the most polysemous word sense\footnote{Currently, we only explore the case of one ambiguous word per sentence, due to the nature of the benchmark. One could extend our approach to multiple ambiguous words by separately sampling examples for each polysemous word and conducting higher-shot prompting - but further research would be needed to find the optimal way to combine these examples. } and search for other occurrences of the same sense in the same development set. Finally, we randomly sample $k$ source-target pairs including such a sense to use as demonstrations in $k$-shot prompting, instead of using random pairs. This technique seemed to return enough examples for our purposes in most cases -- for 5-shot prompting, given a corpus of 1.8M sentences, we observed that we got all 5 matches 92.5\% of the time.

\subsection{Low-rank fine-tuning}
\label{sec:lora-fine-tuning}
Apart from providing relevant examples through prompting, another conventional approach is to optimize the model parameters in a domain adaptation fashion for disambiguation. Considering the computational cost, our work experiments with instruction fine-tuning via low-rank adaptation (LoRA). This technique appends trainable lower-rank decomposition matrices to giant matrices in an LLM that can remain frozen during fine-tuning \citep{lora}. By sacrificing a little performance, this fine-tuning method achieves great parameter efficiency. We aim to adjust LLMs to perform the translation task specifically. In order to maximise an LLM's capability to disambiguate when translating, we follow a careful data selection procedure to identify the most ambiguous sentences in our corpus.

Given the size of LLMs, it would be infeasible to fine-tune them on a large parallel corpus, so we opt to curate a smaller dataset that suits the ambiguous translation task. We would like a balanced mix of sentences with highly polysemous words as well as those with rare senses of a given word. This is to ensure fine-tuning reduces both polysemy degree-related and sense frequency-related biases, as discovered by \citet{campolungo-etal-2022-dibimt} and consequently, maximises disambiguation performance. We, thus, sort our corpora in two ways: one, by the maximum polysemy degree (greatest first) and two, by the minimum sense frequency (rarest first) of all word senses in a given sentence, disambiguated with ESCHER-WSD. We take the top $N/2$ sentences from each set and interleave them to create our final fine-tuning corpus of size $N$. We release our fine-tuning corpus, along with the ESCHER-WSD disambiguation outputs for public use.\footnote{\url{https://data.statmt.org/ambiguous-europarl}}

Once the data is chosen, we follow the fine-tuning paradigm of Alpaca \citep{alpaca}: the model is prompted with an instruction specifying the source and target languages, as well as the test sentence as an input, and the model is expected to respond with the translation.\footnote{\url{https://github.com/tatsu-lab/stanford_alpaca}}

\section{Experiments}
\label{sec:experiments}
In this section, we seek to answer the following research questions:

\begin{enumerate}
    \item \textbf{RQ1:} How do LLMs perform at translation of ambiguous sentences compared to traditional high-performing NMT systems? (Section \ref{sec:dibimt-results})
    \item \textbf{RQ2:} What methods could one use to adapt LLMs for this task and improve performance over naive few-shot prompting? (Section \ref{sec:adapting-llms})
    \item \textbf{RQ3:} How do these disambiguation-adapted LLMs fare in terms of overall translation quality? (Section \ref{sec:overall-mt-results})
\end{enumerate}
\subsection{Models}

To ensure reproducibility, we pick four well-known and high-performing open-source LLMs,\footnote{at the time of experiment formulation} of which we sample seven versions for experimentation:   
\begin{itemize}
    \item BLOOM \citep{bloom}: A fully open-source, multilingual, foundation LLM that supports 46 languages. To establish the range of its capabilities, we explore both the smallest (7.1B) and the largest (176B) versions.
    \item BLOOMZ \citep{muennighoff-etal-2023-crosslingual}: BLOOM instruction-tuned on a multilingual prompting set. Again, we choose the smallest (7.1B) and the largest (176B) versions.
    \item LLaMA \citep{touvron2023llama}: The popular LLM trained by Meta AI, on gigantic datasets ranging up to 1.5T tokens. We evaluate the smallest (7B) and the largest (65B) versions.
    \item Alpaca \cite{alpaca}: A LLaMA model instruction-tuned on a 52K dataset generated using Self-Instruct \citep{wang-etal-2023-self-instruct}.
\end{itemize}

To effectively position these open-source LLMs against traditional NMT systems, we compare them against the best-performing and the most widely used commercial and open-source models:
\begin{enumerate}
    \item DeepL Translator\footnote{\url{https://www.deepl.com/en/translator}}: a SOTA commercial NMT system (accessed on 24th July 2023).
    \item Google Translate\footnote{\url{https://translate.google.com/}}: Probably the most widely used commercial NMT system (accessed on 24th July 2023).
    \item OPUS \citep{tiedemann2020opus}: Small, bilingual, Transformer-based NMT models trained on the OPUS parallel corpora.
    \item mBART50 \citep{tang-etal-2021-multilingual}: Multilingual NMT models pretrained on monolingual corpora from 50 languages, and fine-tuned on the translation task. We report performances of both the English-to-many and many-to-many fine-tuned models.
    \item M2M100 \citep{fan2021beyond}: A massive multilingual NMT model that was trained on 2200 translation directions to support many-to-many translation among 100 languages in total. We compare both the base (418M) and the large (1.2B) versions.
    \item NLLB-200 \citep{costa2022no}: It is the current SOTA in many low-resource pairs, scaling to 200 languages. We experiment with all its variants, where the largest is a mixture-of-experts (MoE) model with 54B parameters. We also benchmark its smaller checkpoints at 1.3B and 3.3B, as well as distilled versions at 0.6B and 1.3B.
\end{enumerate}

We take the results for mBART50, M2M100, and OPUS directly from the DiBiMT leaderboard.\footnote{\url{https://nlp.uniroma1.it/dibimt/public/leaderboard}} We use Hugging Face\footnote{\url{https://huggingface.co/}}
 for accessing and inferencing all other models -- except for Google Translate and DeepL, which are accessed using their respective APIs. Despite their presence on the leaderboard, we re-evaluate these systems since they are being constantly updated.
\begin{table}[!th]
\centering
\begin{tabular}{@{}lcc@{}}
\toprule
\multicolumn{1}{c}{System} & En-Es & En-It \\ \midrule
Similar contexts dev set & 1.81M & 1.73M \\
Fine-tuning corpus & 100K & 100K \\ \bottomrule
\end{tabular}
\caption{Statistics of data used in our experiments, in terms of parallel sentence count.}\vspace{-0.5ex}
\label{tab:data-stats}
\end{table}
\subsection{Experimental setup}
\label{sec:exp-set}

\paragraph{Datasets} In this study, we use the DiBiMT test set for evaluation and measure accuracy across all five translation directions: English to Spanish, Italian, Chinese, Russian, and German, respectively. For validation, we use the development set from FLORES 200 \citep{costa2022no} in our base setting. To search for similar ambiguous contexts (Section \ref{sec:simsentence-prompting}), we require a larger development set to find relevant examples and also to accurately estimate polysemy degree. Hence, we use the Europarl corpus \citep{koehn2005europarl}, disambiguated with ESCHER-WSD. We also use the same disambiguated corpus for fine-tuning, however, we first follow the filtering procedure described in Section \ref{sec:lora-fine-tuning} to create a small corpus full of ambiguous sentences. Validation during fine-tuning is done using 500 randomly sampled sentences from this corpus and the rest is used for training. We detail the data statistics used for these experiments in Table \ref{tab:data-stats}.

\paragraph{LLM prompting setup} Due to memory constraints, and to compare all models fairly, we load LLMs in 8-bit and use a batch size of 1. For generation, we set both beam size and temperature to 1. To prevent repetition in LLM output, we set \texttt{no\_repeat\_ngram\_size} to 4. 
From the LLM's response, we filter out the sentence before the first newline character as the output translation.

\paragraph{LoRA fine-tuning} We inject LoRA modules into all query, key, and value matrices. We set rank to 8, alpha to 8, and dropout to 0.05. For training, we set the effective batch size to 32, the learning rate to 3e-4, and the maximum length to 256. The total training budget is 5 epochs, and we pick the best model checkpoint based on cross-entropy loss on the validation set. The training data is shuffled after every epoch. Inference is done with a beam size of 3, and a maximum generation length of 150.

\begin{table*}[!th]
\small\centering
\begin{tabular}{@{}ccccccccc@{}}
\toprule
System & \# Params & Variant & En-Es & En-It & En-Zh & En-Ru & En-De & Average \\ \midrule \midrule
\multicolumn{9}{c}{\textit{Commercial systems}} \\ \midrule \midrule
DeepL & Unknown & July 2023 & 63.91 &  \textbf{65.47}  & 58.42 & \textbf{67.53} & {\ul \textbf{76.64}} & {\ul \textbf{66.39}} \\
Google Translate & Unknown & July 2023 & 54.73 & 53.59 & 52.09 & 62.03 & \textbf{\textit{67.35}} & 57.96 \\ \midrule \midrule
\multicolumn{9}{c}{\textit{Open-source NMT systems}} \\ 
\midrule 
\midrule 
OPUS & 74M & Bilingual En-X models & 36.79 & 29.93 & 25.94 & 28.71 & 27.04 & 29.68  \\ 
 [0.5ex] \cdashline{1-9}\noalign{\vskip 0.5ex}
\multirow{2}{*}{mBART50 } & 611M & One-to-Many & 31.31 & 26.62 & 26.63 & 30.93 & 26.43 & 28.38 \\
 & 611M & Many-to-Many & 29.98 & 25.89 & 28.12 & 27.54 & 24.25 & 27.16  \\
 [0.5ex] \cdashline{1-9}\noalign{\vskip 0.5ex}
\multirow{2}{*}{M2M100 } & 418M & Base & 22.35 & 17.27 & 12.34 & 17.01 & 15.62 & 16.92 \\
 & 1.2B & Large & 28.81 & 23.16 & 17.30 & 27.03 & 22.87 & 23.83  \\ 
 [0.5ex] \cdashline{1-9}\noalign{\vskip 0.5ex}
\multirow{5}{*}{NLLB-200 } & 0.6B & Distilled version & 40.93 & 36.38 & 28.64 & 47.13 & 33.41 & 37.30 \\
 & 1.3B & Distilled version & 50.40 & 53.65 & 41.15 & 54.52 & 52.81 & 50.51 \\
 & 1.3B & Original checkpoint & 48.81 & 48.43 & 37.31 & 54.36 & 48.93 & 47.57 \\
 & 3.3B & Original checkpoint & 53.23 & 57.23 & 39.95  & 57.44 & 56.24 & 52.82 \\
 & 54B & Mixture of Experts & 61.33 & {\ul \textbf{67.19}} & 48.02  & {\ul \textbf{67.88}} & \textbf{67.97} & \textbf{62.48} \\ \midrule \midrule
\multicolumn{9}{c}{\textit{LLaMA family LLMs}} \\ \midrule \midrule
\multirow{6}{*}{LLaMA} & \multirow{3}{*}{7B} & 1-shot prompting & 53.64 & 48.84 & 30.61$^\dag$ & 60.65 & 57.41 & 50.23 \\
 &  & 3-shot prompting & 55.53 & 50.53 & 30.52$^\dag$ & 57.31 & 55.34 & 49.85 \\
 &  & 5-shot prompting & 56.33 & 48.66 & 27.92$^\dag$ & 56.83 & 55.26 & 49.00 \\ [0.5ex] \cdashline{2-9}\noalign{\vskip 0.5ex}
 & \multirow{3}{*}{65B} & 1-shot prompting & 56.57 & 60.22 & 44.73$^\dag$ & 65.71 & 62.05 & 57.86 \\ 
 &  & 3-shot prompting & 59.83 & 60.18 & 42.77$^\dag$ & \textbf{\textit{67.45}} & 63.41 & 58.73 \\
 &  & 5-shot prompting & 60.78 & \textbf{\textit{63.47}}  & 42.49$^\dag$ & 66.31 & 62.98 & \textbf{\textit{59.21}}  \\ 
 [0.5ex] \cdashline{1-9}\noalign{\vskip 0.5ex}
Alpaca  & 7B & 0-shot prompting & 49.75 & 45.24 & 29.63$^\dag$ & 55.23 & 51.52 & 46.27 \\ \midrule \midrule
\multicolumn{9}{c}{\textit{BLOOM family LLMs}} \\ \midrule \midrule
\multirow{4}{*}{BLOOM } & 7.1B & 1-shot prompting & 55.69 & 28.79$^\dag$ & 51.08 & 40.00$^\dag$ & 29.67$^\dag$ & 41.05 \\ [0.5ex] \cdashline{2-9}\noalign{\vskip 0.5ex}
 & \multirow{3}{*}{176B} & 1-shot prompting & 63.66 & 42.02$^\dag$ & 60.30 & 43.22$^\dag$ & 37.04$^\dag$ & 49.25 \\
 &  & 3-shot prompting & 64.52 & 46.33$^\dag$ & 61.20 & 44.30$^\dag$ & 36.69$^\dag$ & 50.61 \\
 &  & 5-shot prompting & 65.53 & 45.99$^\dag$ & 61.73 & 42.92$^\dag$ & 38.06$^\dag$ & 50.85  \\ 
 [0.5ex] \cdashline{1-9}\noalign{\vskip 0.5ex}
\multirow{6}{*}{BLOOMZ } & \multirow{2}{*}{7.1B} & 0-shot prompting & 56.89 & 33.91$^\dag$ & 53.2 & 33.33$^\dag$ & 21.67$^\dag$ & 39.80 \\ 
& & 1-shot prompting & 60.87 & 40.68$^\dag$ & 52.37 & 33.33$^\dag$ & 30.65$^\dag$ & 43.58 \\ [0.5ex] \cdashline{2-9}\noalign{\vskip 0.5ex}
 & \multirow{4}{*}{176B} & 0-shot prompting & 62.67 & 45.78$^\dag$ & 61.87 & 47.98$^\dag$ & 44.06$^\dag$ & 52.47 \\
 & & 1-shot prompting & \textbf{\textit{64.35}} & 49.31$^\dag$ & {\ul \textbf{66.57}} & 51.88$^\dag$ & 43.92$^\dag$ & 55.21 \\
 &  & 3-shot prompting & \textbf{67.31} & 45.91$^\dag$ & \textbf{64.44} & 53.42$^\dag$ & 45.08$^\dag$ & 55.23 \\
 &  & 5-shot prompting & {\ul \textbf{68.55}} & 49.22$^\dag$ & \textbf{\textit{63.36}} & 52.60$^\dag$ & 44.94$^\dag$ & 55.73 \\ \bottomrule
\end{tabular}
\caption{Accuracies on DiBiMT test for establish NMT systems and LLMs, using naive $k$-shot prompting. For Alpaca, we can only use 0-shot prompting due to its particular prompt template. We highlight the top three scores per language in bold, with the best underlined as well, the 2nd best as is, and the 3rd best italicized. We indicate scores for unseen languages (ie. not intentionally included in pretraining) with a $\dag$.}
\vspace{-0.15in}
\label{tab:dibimt-results}
\end{table*}

\subsection{LLMs vs NMT systems on DiBiMT}
\label{sec:dibimt-results}

We show the results of our experiments in Table \ref{tab:dibimt-results}. For the purposes of the subsequent discussion, we note here that LLaMA was not intentionally trained on Chinese and is, thus, an `unseen' language. Similarly, for BLOOM, Chinese and Spanish are ``seen'' and the rest are ``unseen''. We share our key observations below:

\begin{enumerate}
\item \textbf{LLMs usually match or beat massive MT models on seen languages.} Except for the very rich-resourced En-De, where supervised MT systems appear to have an edge, LLaMA 65B mostly matches the SOTA NMT systems (namely DeepL and NLLB-200). Furthermore, BLOOMZ sets a new SOTA in its seen languages, Spanish and Chinese, and outperforms DeepL by margins of 7.3\% and 12.2\% respectively. These improvements against such strong, supervised massive NMT systems are particularly remarkable since our corresponding setup for inferencing the LLMs is quite cheap -- as we noted previously, this is only naive few-shot prompting of an 8-bit quantized model, with a beam size of 1.

    \item \textbf{LLMs perform relatively worse for unseen languages, but they can still be much better than some supervised MT models.} We note that relative to seen languages, LLaMA underperforms in translation to Chinese. Similarly, BLOOM performs worse for its' unseen languages of German, Italian, and Russian. Still, LLMs yield reasonable performance here that is still much better than some supervised NMT systems. For example, BLOOMZ-7B achieves 40.68\% accuracy in English-Italian,  which is about 35.9\% more than OPUS, 52.8\% more than mBART50 and 75\% more than M2M100-1.2B. While NLLB-200 does outperform BLOOMZ-7B, our results just highlight the power of pretraining at scale.

    \item \textbf{Scale helps improve performance for ambiguity translation}. Continuing from the last point, similar to NMT models that improve with scale (e.g. NLLB-200), we observe that LLMs too perform consistently better at ambiguous translation on scaling up to their larger variants. This applies to the translation of both seen and unseen languages. That said, the lighter models, such as LLaMA 7B or BLOOM 7B, also perform quite well and in many cases, 1-shot prompting of these LLMs is almost as good as NLLB translations.
    
    \item \textbf{LLM performance does improve on average with more demonstrations, but this is not uniform.} On average, we observe that 5-shot prompting works best, followed by 3-shot and then 1-shot, though some outliers exist for LLaMA 7B. Moreover, when looking at the performance of individual language pairs, we note that the improvement trend is not uniform, and it is possible a 3-shot translation outperforms a 5-shot one. This aligns with the finding of \citet{zhang2023prompting}, who reach the same conclusion regarding overall MT quality. Nonetheless, as we show in Section \ref{sec:better-icl}, accuracy does significantly improve when we provide relevant and helpful examples -- suggesting quality of demonstrations matters more than quantity.
    
\item \textbf{General-purpose instruction-tuned LLMs consistently outperform foundation LLMs.} Interestingly, we observe that 1-shot prompting of a general-purpose instruction-tuned LLM like BLOOMZ often significantly outperforms 5-shot prompting of BLOOM, even on the very specific task of ambiguity translation. In fact, even with 0-shot prompting, models like Alpaca 7B, BLOOMZ 7B and BLOOMZ 176B perform reasonably well, matching some supervised MT systems. We observed that this did not work for foundation LLMs like BLOOM 165B and LLaMA 7B, and 0-shot prompting of these models yielded hallucinations in many cases.

\end{enumerate}

Lastly, we include a qualitative comparison of DeepL and BLOOMZ 176B translations for the En-Zh pair in the Appendix (see Table \ref{tab:manual-en-zh}) -- where we observe that BLOOMZ generates more contextual translations, relatively speaking, while its counterpart tends to translate literally in many cases.
\subsection{Adapting LLMs for ambiguous MT}
\label{sec:adapting-llms}

This section reports experiments with two proposed strategies to enable LLMs to disambiguate better and improve performance on the ambiguous translation task. While both methods are shown to significantly improve performance, we include a discussion of the relative tradeoffs between the techniques in Appendix \ref{sec:tradeoffs}.

\subsubsection{Improving In-Context Learning by leveraging similar ambiguous contexts}
\label{sec:better-icl}

\begin{table*}[thb]
\begin{subtable}[t]{0.48\textwidth}
\small\centering
\setlength{\tabcolsep}{0.8ex}
\begin{tabular}{lcccccc}
\toprule
\multicolumn{1}{c}{\multirow{2}{*}{System}} & \multicolumn{2}{c}{1-shot} & \multicolumn{2}{c}{3-shot} & \multicolumn{2}{c}{5-shot} \\
\cmidrule(lr){2-3}\cmidrule(lr){4-5}\cmidrule(lr){6-7} 
 & Rand. & Sim. & Rand. & Sim. & Rand. & Sim. \\ \midrule
DeepL & \multicolumn{6}{c}{{---}63.91{---}} \\
NLLB-200 54B & \multicolumn{6}{c}{---\textit{\textbf{61.33}}---} \\ 
 [0.3ex] \cdashline{1-7}\noalign{\vskip 0.5ex}
LLaMA 7B & 53.64 & \textbf{54.01} & \textbf{55.53} & 52.52 & \textbf{56.33} & 54.45 \\
LLaMA 65B & 56.57 & \textbf{59.38} & 59.83 & \textbf{62.44} & 60.78 & \textbf{63.74} \\
BLOOM 176B & \textbf{63.66} & 62.44 & 64.52 & \textbf{66.19} & 65.53 & \textbf{68.22} \\
BLOOMZ 176B & 64.35 & \textbf{69.57} & 67.31 & \textbf{71.15} & 68.55 & {\ul \textbf{71.33}} \\
\bottomrule
\end{tabular}
\caption{English-Spanish}\label{tab:1a}
\end{subtable}
\hfill
\begin{subtable}[t]{0.48\textwidth}
\small\centering
\setlength{\tabcolsep}{0.8ex}
\begin{tabular}{lcccccc}
\toprule
\multicolumn{1}{c}{\multirow{2}{*}{System}} & \multicolumn{2}{c}{1-shot} & \multicolumn{2}{c}{3-shot} & \multicolumn{2}{c}{5-shot} \\
\cmidrule(lr){2-3}\cmidrule(lr){4-5}\cmidrule(lr){6-7}
 & Rand. & Sim. & Rand. & Sim. & Rand. & Sim. \\ \midrule
DeepL & \multicolumn{6}{c}{{---}65.47{---}} \\
NLLB-200 54B & \multicolumn{6}{c}{---\textit{\textbf{67.19}}---} \\ 
 [0.3ex] \cdashline{1-7}\noalign{\vskip 0.5ex}
LLaMA 7B & 48.84 & \textbf{49.47} & 50.53 & \textbf{53.85} & 48.66 & \textbf{52.17} \\
LLaMA 65B & \textbf{60.22} & 59.77 & 60.18 & \textbf{64.94} & 63.47 & {\ul \textbf{65.33}} \\
BLOOM 176B & 42.02 & \textbf{43.17} & 46.33 & \textbf{48.09} & 45.99 & \textbf{50.00} \\
BLOOMZ 176B & 49.31 & \textbf{49.60} & 45.91 & \textbf{50.73} & 49.22 & \textbf{50.53} \\
\bottomrule
\end{tabular}

\caption{English-Italian}\label{tab:enit_simsents}
\end{subtable}

\caption{1-shot, 3-shot and 5-shot results for En-Es and En-It prompting with randomised examples (\texttt{Rand.}) versus similar contexts (\texttt{Sim.}). The best-performing systems from Table \ref{tab:dibimt-results}, i.e. DeepL and NLLB-200 are chosen as baselines. For LLMs, for each setting, the better-performing baseline between \texttt{Rand.} and \texttt{Sim.} is highlighted in bold. The overall best score (among all LLMs) is underlined as well, while the best NMT system is also italicized. }
\label{tab:simsentence-results}
\end{table*}

Rather than selecting our examples randomly as in our naive setting, we employ the data selection procedure described in Section \ref{sec:simsentence-prompting} to discover other examples that contain the same word sense as the most polysemous sense in the input sentence. We report our scores in Table \ref{tab:simsentence-results}, and our findings below:

\begin{enumerate}
    \item \textbf{Similar contexts yield more improvements as the example count increases} We observe that for 1-shot prompting, similar contexts perform comparably or slightly better than random examples. However, the gains increase substantially as we move towards 3-shot and 5-shot prompting. We can understand this from the intuition that 1-shot prompting likely just guides the LLM towards generating a reasonable translation, whereas with more relevant examples, it learns to disambiguate better and translate in context accordingly.

    \item \textbf{Larger models observe greater and more consistent gains than smaller LLMs} Compared to LLaMA 7B, the other LLMs (LLaMA 65B, BLOOM 176B and BLOOMZ 176B) yield much larger accuracy improvements on a more uniform basis. This is probably because scaling up allows LLMs to model polysemous words better in their semantic space, facilitating effective in-context learning of disambiguation capabilities. 
\end{enumerate}

\subsubsection{Fine-tuning with ambiguous corpora}
\label{sec:ft-ambiguous-corpora}

\begin{table*}[!th]
\centering
\resizebox{0.75 \textwidth}{!}{\begin{tabular}{@{}ccccccc@{}}
\toprule
\multirow{2}{*}{System} & \multicolumn{3}{c}{En-Es} & \multicolumn{3}{c}{En-It} \\
\cmidrule(lr){2-4}\cmidrule(lr){5-7}
 & Alpaca 7B & BLOOM 7B & BLOOMZ 7B & Alpaca 7B & BLOOM 7B & BLOOMZ 7B \\ \midrule
w/o FT & 49.75 & 55.69 & 60.87 & 45.24 & 28.79 & 40.68 \\
FT (Best Loss) & 63.27 & 57.86 & 60.39 & 59.62 & 37.72 & 39.73 \\
FT (Best Acc.) & 63.31 & 59.72 & 61.56 & 59.77 & 42.40 & 44.73 \\ \bottomrule
\end{tabular}}
\caption{DiBiMT Accuracies after fine-tuning Alpaca 7B, BLOOM 7B, and BLOOMZ 7B on En-Es and En-It pairs. The ``Best Loss'' baseline refers to the checkpoint with best cross-entropy loss on the validation set. The ``Best Acc.'' baseline refers to the checkpoint with best DiBiMT accuracy, on evaluating the last checkpoint after each epoch.}
\label{tab:ft-results}
\end{table*}

\begin{figure*}[!th]
\centering
   \begin{subfigure}{0.49\linewidth} \centering
     \includegraphics[scale=0.4]{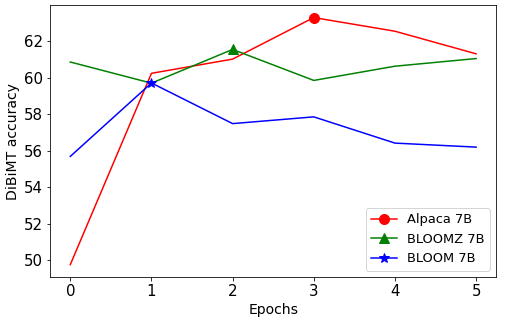}
     \caption{English-Spanish}\label{fig:enes_ft}
   \end{subfigure}
   \begin{subfigure}{0.49\linewidth} \centering
     \includegraphics[scale=0.4]{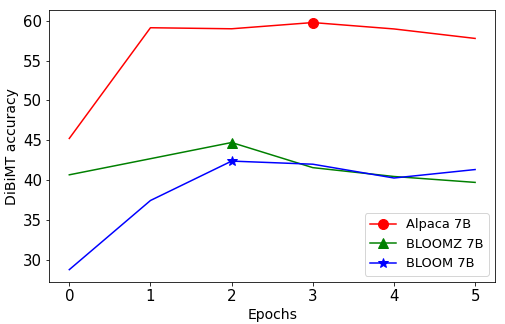}
     \caption{English-Italian}\label{fig:enit_ft}
   \end{subfigure}
\caption{DiBiMT accuracy at the end of every epoch, for the LoRA fine-tuned LLMs} \label{fig:accuracy_epoch_curve}
\end{figure*}

\begin{figure*}[!th]
\centering
   \begin{subfigure}{0.49\linewidth} \centering
     \includegraphics[scale=0.4]{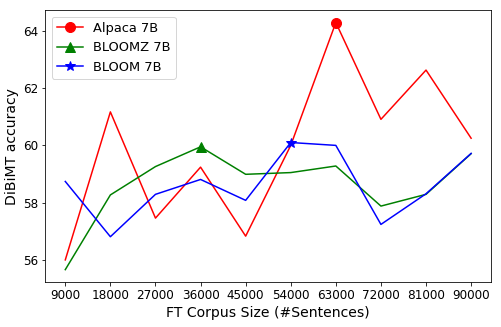}
     \caption{English-Spanish}\label{fig:enes_ft_epoch1}
   \end{subfigure}
   \begin{subfigure}{0.49\linewidth} \centering
     \includegraphics[scale=0.4]{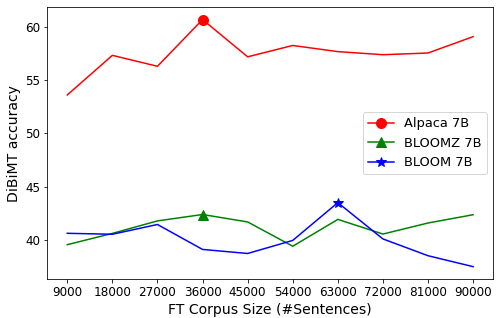}
     \caption{English-Italian}\label{fig:enit_ft_epoch1}
   \end{subfigure}
\caption{DiBiMT accuracy vs fine-tuning (FT) corpus size in terms of parallel sentence count. These results are obtained from evaluating checkpoints at every 300 steps in the 1st epoch - which roughly corresponds to about 9K sentences, since we use a batch size of 32.} \label{fig:accuracy_epoch1_curve}
\end{figure*}

We fine-tune Alpaca 7B, BLOOM 7B and BLOOMZ 7B in En-Es and En-It directions using the data described in Section \ref{sec:exp-set}. We show our results when prompting these fine-tuned LLMs in Table \ref{tab:ft-results}. We make the following observations:

\begin{enumerate}
    \item \textbf{Fine-tuning generally improves performance.} We observe that fine-tuned LLMs significantly outperform their non-finetuned versions in most cases. The biggest improvement is observed for BLOOM 7B in En-It, where accuracy increases by as high as 47.73\%, indicating the effectiveness of our method. The only exception to this is when the LLM is already strong, such as BLOOMZ 7B at En-Es, and then the improvements are marginal. But even so, strong instruction-tuned LLMs like BLOOMZ still gain significantly from fine-tuning on the En-It pair -- where it was originally weaker due to Italian being an unseen language during pretraining.

    \item \textbf{Fine-tuning for 2-3 epochs is sufficient.} We plot the DiBiMT accuracy versus epoch curves in Figure  \ref{fig:accuracy_epoch_curve} where the performance is evaluated after each epoch. We observe that in all cases, accuracy peaks between the 1st and the 3rd epoch, after which it mostly plateaus or dips slightly - suggesting that one does not need to fine-tune these LLMs for too long.

    \item \textbf{Fine-tuning improves LLM performance until about 65K training samples.} We now try to answer the Research Question of how many training samples we need for fine-tuning these LLMs, to get optimal performance. We plot the Accuracy vs corpus size graph in Figure \ref{fig:accuracy_epoch1_curve}, where we indicate corpus size by the number of parallel sentences. We observe that accuracy increases non-monotonically with an increase in corpus size, but peaks anywhere between 36K-63K training samples, which seems to depend on the pre-existing capabilities of the LLM. For a raw foundation LLM like BLOOM 7B, relatively more fine-tuning data (54K-63K) appears to be beneficial. Alpaca 7B, which has been instruction-tuned on an English-only dataset, also seems to benefit from further fine-tuning---especially for En-Es, accuracy peaks after 63K training samples. However, for a powerful LLM like BLOOMZ that has been instruction-tuned on a large multilingual dataset like xP3 \citep{muennighoff-etal-2023-crosslingual}, fine-tuning on smaller datasets (at most 36K sentences, in our case) appears to suffice.
\end{enumerate}

\subsection{Overall MT performance of disambiguation-adapted LLMs}
\label{sec:overall-mt-results}
\begin{table*}[htbp]
\centering\small
\begin{tabular}{@{}lcccccc@{}}
\toprule
\multicolumn{1}{c}{\multirow{2}{*}{System}} & \multicolumn{3}{c}{En-Es} & \multicolumn{3}{c}{En-It} \\ 
\cmidrule(lr){2-4}\cmidrule(lr){5-7}
 & spBLEU & chrF++ & COMET22 & spBLEU & chrF++ & COMET22 \\ \midrule
NLLB-200 54B & 32.50 & 53.79 & 0.86 & 37.60 & 57.33 & 0.89 \\ 
[0.3ex] \cdashline{1-7}\noalign{\vskip 0.5ex}
Alpaca 7B (0-shot) & 23.90 & 47.30 & 0.83 & 23.30 & 46.40 & 0.83 \\
LLaMA 7B (1-shot) & 23.20 & 46.20 & 0.82 & 22.10 & 45.00 & 0.82 \\
LLaMA 65B (1-shot) & 27.20 & 49.70 & 0.83 & 28.50 & 50.50 & 0.85 \\
BLOOM 7B (1-shot) & 24.00 & 46.30 & 0.82 & \phantom{$^\dag$}10.00$^\dag$ & \phantom{$^\dag$}33.40$^\dag$ & \phantom{$^\dag$}0.63$^\dag$ \\
BLOOM 176B (1-shot) & 28.60 & 51.20 & 0.85 & \phantom{$^\dag$}20.80$^\dag$ & \phantom{$^\dag$}45.20$^\dag$ & \phantom{$^\dag$}0.81$^\dag$ \\ 
[0.3ex] \cdashline{1-7}\noalign{\vskip 0.5ex}
Alpaca 7B (FT, 0-shot) & 27.40 & 50.20 & 0.85 & 29.20 & 51.40 & 0.87 \\
BLOOM 7B (FT, 0-shot) & 28.70 & 51.00 & 0.86 & 20.90 & 45.80 & 0.80 \\
\bottomrule
\end{tabular}

\vspace*{3ex}

\centering\small
\setlength{\tabcolsep}{1ex}
\begin{tabular}{@{}lccccccccc@{}}
\toprule
\multicolumn{1}{c}{\multirow{2}{*}{System}} & \multicolumn{3}{c}{En-Zh} & \multicolumn{3}{c}{En-Ru} & \multicolumn{3}{c}{En-De} \\
\cmidrule(lr){2-4}\cmidrule(lr){5-7}\cmidrule(lr){8-10}
 & spBLEU & chrF++ & COMET22 & spBLEU & chrF++ & COMET22 & spBLEU & chrF++ & COMET22 \\ \midrule
NLLB-200 54B & 23.10\phantom{$^\dag$} & 22.83\phantom{$^\dag$} & 
0.82\phantom{$^\dag$} & 38.00\phantom{$^\dag$} & 56.34\phantom{$^\dag$} & 0.90\phantom{$^\dag$} & 44.80\phantom{$^\dag$} & 62.79\phantom{$^\dag$} & 0.88\phantom{$^\dag$} \\
[0.3ex] \cdashline{1-10}\noalign{\vskip 0.5ex}
Alpaca 7B (0-shot) & \phantom{0}4.80$^\dag$ & 10.40$^\dag$ & 0.62$^\dag$ & 21.80\phantom{$^\dag$} & 42.60\phantom{$^\dag$} & 0.82\phantom{$^\dag$} & 27.30\phantom{$^\dag$} & 50.30\phantom{$^\dag$} & 0.82\phantom{$^\dag$} \\
LLaMA 7B (1-shot) & \phantom{0}5.60$^\dag$ & 10.80$^\dag$ & 0.66$^\dag$ & 20.70\phantom{$^\dag$} & 41.20\phantom{$^\dag$} & 0.79\phantom{$^\dag$} & 22.80\phantom{$^\dag$} & 45.40\phantom{$^\dag$} & 0.78\phantom{$^\dag$} \\
LLaMA 65B (1-shot) & 13.80$^\dag$ & 17.60$^\dag$ & 0.77$^\dag$ & 26.70\phantom{$^\dag$} & 46.10\phantom{$^\dag$} & 0.82\phantom{$^\dag$} & 31.80\phantom{$^\dag$} & 52.80\phantom{$^\dag$} & 0.81\phantom{$^\dag$} \\
BLOOM 7B (1-shot) & 19.00\phantom{$^\dag$} & 19.50\phantom{$^\dag$} & 0.83\phantom{$^\dag$} & \phantom{0}3.70$^\dag$ & 22.30$^\dag$ & 0.46$^\dag$ & \phantom{0}8.20$^\dag$ & 31.70$^\dag$ & 0.51$^\dag$ \\
BLOOM 176B (1-shot) & 25.10\phantom{$^\dag$} & 23.80\phantom{$^\dag$} & 0.86\phantom{$^\dag$} & 10.30$^\dag$ & 31.80$^\dag$ & 0.65$^\dag$ & 19.90$^\dag$ & 45.40$^\dag$ & 0.74$^\dag$ \\
\bottomrule
\end{tabular}

\caption{FLORES 200 results for $k$-shot prompting of some key LLMs used in this work, compared with the NLLB-200 baseline. We also include results for the LoRA fine-tuned models, for the En-Es and En-It pairs. Same as the previous notation, we indicate all unseen language results with a $^\dag$. We observe similar trends in all standard MT metrics, as those observed with DiBiMT accuracy.}
\vspace{-0.1in}
\label{tab:flores200-results}
\end{table*}

Lastly, for completeness, we also evaluate the overall translation quality of the key LLMs used in this work -- since we are interested in noting how well the reported disambiguation accuracies extend to overall MT performance. While choosing our test set, we want to ensure it is recently released (ideally within the last year) to minimize the chances of its inclusion in the pretraining corpora of LLMs. We, thus, choose FLORES 200 \citep{costa2022no} as our test set since it satisfies this criterion and also supports all our languages of evaluation. We use spBLEU\footnote{\scriptsize \texttt{nrefs:1|case:mixed|eff:no|tok:flores101|smooth:exp|version:2.3.1}} \citep{goyal2022flores}, chrF++\footnote{\scriptsize \texttt{nrefs:1|case:mixed|eff:yes|nc:6|nw:2|space:no|version:2.3.1}} \citep{popovic2017chrf++} and COMET22 \citep{rei-etal-2022-comet} using the \texttt{wmt22-comet-da} model as metrics. In this setting, we evaluate Alpaca with 0-shot prompting, while LLaMA 7B, LLaMA 65B and BLOOM 176B use the 1-shot setup. NLLB-200 is our primary supervised NMT baseline. We also evaluate LoRA fine-tuned versions of Alpaca 7B and BLOOM 7B, from section \ref{sec:ft-ambiguous-corpora}, on the English-Spanish and English-Italian pairs. We exclude BLOOMZ from this evaluation since it is instruction-tuned on FLORES200. We report our results in Table \ref{tab:flores200-results}.

\begin{table}[htbp]
\centering\small
\setlength{\tabcolsep}{2.5ex}
\begin{tabular}{cccc}
\toprule
 & \makecell{spBLEU\\w/ acc.} & \makecell{ChrF++\\w/ acc.} & \makecell{COMET22\\w/ acc.} \\ \midrule
$\rho$ & 0.83 & 0.56 & 0.76 \\
$p$-value & 0.0001 & 0.0039 & 0.0010 \\ \bottomrule
\end{tabular}
\caption{Pearson's correlation $\rho$ \citep{benesty2009pearson} between DiBiMT accuracy and spBLEU, chrF++, and COMET22 respectively, together with p-values.}
\vspace{-0.1in}
\label{tab:correlation-table}
\end{table}

We observe trends similar to those of our DiBiMT experiments. BLOOM 176B performs well in translation of seen languages, performing comparably to NLLB-200 in English-Spanish and outperforming it in English-Chinese. This is particularly the case for COMET22 scores, a metric which has shown high correlations with human evaluation, ranking second in the WMT22 Metrics shared task \citep{freitag-etal-2022-results}. For the other languages, LLaMA 65B usually performs better than BLOOMZ, but in the 1-shot prompting setup, it is unable to beat the NLLB-200 54B MOE. We also notice that the fine-tuned versions of Alpaca 7B and BLOOM 7B consistently outperform their vanilla counterparts -- suggesting our techniques to improve disambiguation performance also boost overall translation quality.

Thus, while we evaluate some key LLMs to verify consistent trends, we want to avoid re-running all our baselines on FLORES200. So, we try to answer a broader question: how well does disambiguation accuracy on DiBiMT correlate with standard MT metrics? We conduct a Pearson's correlation test \citep{benesty2009pearson} between the accuracy metric and spBLEU, chrF++, and COMET22 respectively. We report our results in Table \ref{tab:correlation-table}, and find that all MT quality metrics correlate positively with accuracy---with $p$-values of the two-sided alternative hypothesis being much lesser than 0.05 in all cases. We discover that spBLEU and COMET22 exhibit higher correlations than chrF++. We hypothesize that this could be due to the character-level chrF++ being less sensitive to word-level senses. Overall, the results of Tables \ref{tab:flores200-results} and \ref{tab:correlation-table} suggest that the significant accuracy improvements noted earlier are not at the cost of translation quality, and in turn, could yield improvements in overall MT scores too.

\section{Conclusion}
\label{sec:conclusion}
In this work, we studied the capabilities of LLMs to handle ambiguity during machine translation. We choose seven of the most widely used foundation and instruction-tuned LLMs and compare accuracy with SOTA commercial and open-source NMT systems on the DiBiMT translation benchmark. Out of 5 language directions, we report scores comparable to the SOTA on two (En-Ru, En-It) and set a new SOTA on two others (En-Zh, En-Es). We then present two techniques that significantly improve disambiguation accuracy: in-context learning with similar contexts, and fine-tuning on an ambiguous corpus. We end the paper with an evaluation of overall MT quality. We hope the methods and findings shared in this work could guide future researchers studying ambiguity in translation.

\newpage
\section*{Limitations}
\label{sec:limitations}
In this work, we attempt to note overall trends in LLM performance as compared to conventional NMT systems and, based on our results, suggest methods that generally improve performance. That said, there are exceptions to these trends - prompting with similar contexts can, at times, degrade performance and so can increasing the number of demonstrations (see Table \ref{tab:simsentence-results}). But there is some consistency here too that these observations mostly apply to smaller LLMs (such as LLaMA 7B) while the larger LLMs benefit more significantly. Also, as noted in Section \ref{sec:better-icl}, in a small percentage of cases (7.5\%), we are unable to find 5 matches when attempting 5-shot prompting with similar contexts. In such cases, it might be worthwhile, from a performance perspective, to use random demonstrations; nonetheless, since we are interested in verifying the utility of similar contexts and also since there are only a few cases where it might be pertinent, we do not explore this.

\section*{Acknowledgements}
This work has received funding from UK Research and Innovation under the UK government's Horizon Europe funding guarantee [grant numbers 10039436 and 10052546].

The computations described in this research were performed using the Baskerville Tier 2 HPC service (https://www.baskerville.ac.uk/). Baskerville was funded by the EPSRC and UKRI through the World Class Labs scheme (EP/T022221/1) and the Digital Research Infrastructure programme (EP/W032244/1) and is operated by Advanced Research Computing at the University of Birmingham.


\bibliography{anthology,custom}
\bibliographystyle{acl_natbib}

\appendix

\begin{table*}
    \centering\small
    \begin{tabular}{lll}
        \toprule
        Source & \multicolumn{2}{l}{Tap the \hlblue{head} of the drum for this roll.} \\
        DeepL & \chinese{敲击鼓的\hlred{头部}进行滚奏。} & \textit{head, literal} \\
        BLOOMZ & \chinese{敲击\hlgreen{鼓面}，发出这个鼓点。} & \textit{striking surface of a drum} \\ 
        \midrule
        Source & \multicolumn{2}{l}{they tracked him back toward the \hlblue{head} of the stream.} \\
        DeepL & \chinese{他们跟踪他回到溪\hlred{头} .} & \textit{head (literal and unnatural)} \\ 
        BLOOMZ & \chinese{他们跟着他回到了河的\hlgreen{上游}。} & \textit{upstream} \\
        \midrule
        Source & \multicolumn{2}{l}{The expedition followed the river all the way to the \hlblue{head}.} \\
        DeepL & \chinese{探险队沿着河流一直走到河\hlred{头}。} & \textit{head (literal and unnatural)} \\
        BLOOMZ & \chinese{探险队顺着河一直走到\hlgreen{源头}。} & \textit{source} \\ 
        \midrule
        Source & \multicolumn{2}{l}{How much \hlblue{head} do you have at the Glens Falls feeder dam?} \\
        DeepL & \chinese{格伦瀑布支坝的\hlgreen{水头}有多大？} & \textit{hydraulic head} \\
        BLOOMZ & \chinese{你有多少\hlred{头}牛在格伦瀑布的蓄水池里？} & \textit{(a classifier word to express quantities (of cows))} \\ 
        \midrule
        Source & \multicolumn{2}{l}{the office was full of secret \hlblue{heads}.} \\
        DeepL & \chinese{办公室里到处都是秘密\hlgreen{头目} .} & \textit{leader, ringleader} \\
        BLOOMZ & \chinese{办公室挤满了神秘的\hlgreen{首脑}。} & \textit{leader, head of state} \\ 
        \bottomrule
    \end{tabular}
    \caption{Manual inspection on English-to-Chinese translation focused on the disambiguation of ``head'', corresponding to the first five test instances in DiBiMT. The baselines are DeepL and BLOOMZ 176B, the highest performing NMT system and LLM for this pair (from Table \ref{tab:dibimt-results}). The reported annotations are obtained from a native Chinese speaker who was invited to label the sense of the translated ambiguous word.}
    \vspace{-0.08in}
    \label{tab:manual-en-zh}
\end{table*}

\newpage
\section{Appendix}
\label{sec:appendix}

\subsection{Qualitative comparison: BLOOMZ vs DeepL}

We choose the best-performing LLM and the SOTA MT system from Table \ref{tab:dibimt-results} -- focusing on the En-Zh pair since LLMs seem to yield the highest gains there. With the help of a native Chinese speaker, we got hypotheses from these two systems annotated, for the first 5 sentences of the DiBiMT test set. We observe that although there are cases where DeepL gets it right over BLOOMZ (example 4) or where both are correct (Example 5), in many instances BLOOMZ appears to generate more contextual (and less literal) translations. We hypothesize that this could potentially be due to the former's powerful language modelling abilities

\subsection{Trade-off between prompting and fine-tuning}
\label{sec:tradeoffs}

We show in Section \ref{sec:adapting-llms} that both prompting with similar contexts through In-Context Learning (ICL) and LoRA fine-tuning can significantly improve performance. However, depending on the use case, it might be better to favour one over the other. For instance, in production environments, LLMs that are LoRA fine-tuned on ambiguous text can provide powerful disambiguation performance, while also being more feasible to deploy and run at scale. In contrast, ICL with $k$-shot prompting, especially for higher values of $k$, can significantly increase query size and memory consumption, necessitating reduced batch size and thus, throughput. 

However, conducting ICL with similar ambiguous contexts can be used to query LLMs as large as LLaMA 65B and BLOOMZ 176B and yield performance comparable to SOTA MT systems (see Table \ref{tab:simsentence-results}). The preprocessing cost overhead of such a method, namely disambiguating the test set, is also low - it took us about 13 seconds to disambiguate a test set of about 500 sentences on 1 Nvidia GeForce RTX 3090. In contrast, the one-time cost of fine-tuning can be quite expensive---for instance, it took us 44 hours to fine-tune an Alpaca 7B with LoRA on a single Nvidia Tesla A100 40G. Thus, in GPU-scarce settings where the costs of LoRA fine-tuning are prohibitive, it might be favourable to use ICL to query massive LLMs and obtain SOTA performances. In contrast, production environments are likely to prefer the fine-tuned LLMs, since the one-off fine-tuning costs can be amortized.

\end{document}